\documentclass{article} 
\usepackage{iclr2023_conference_tinypaper,times}

\usepackage{hyperref}
\usepackage{url}
\usepackage{tabularx}

\usepackage{graphicx}
\usepackage{arydshln}

\title{Frustratingly Simple Prompting-based Text Denoising}
\author{Jungyeul Park\thanks{
This work was supported in part by Oracle Cloud credits and related resources provided by Oracle for Research.
} \\
Department of Linguistics\\
The University of British Columbia\\
Vancouver, BC V6T 1Z4, Canada \\
\texttt{jungyeul@mail.ubc.ca}  \\ \And
Mengyang Qiu\\
Department of Psychology\\
Trent University\\
Peterborough, ON K9L 0G2, Canada \\
\texttt{mengyangqiu@trentu.ca}}

\iclrfinalcopy 
\begin{document}

\maketitle

\begin{abstract}
This paper introduces a novel perspective on the automated essay scoring (AES) task, challenging the conventional view of the ASAP dataset as a static entity. Employing simple text denoising techniques using prompting, we explore the dynamic potential within the dataset. 
While acknowledging the previous emphasis on building regression systems, our paper underscores how making minor changes to a dataset through text denoising can enhance the final results. 
\end{abstract}

\section{Introduction}

Text denoising is a crucial step in natural language processing (NLP) and text analysis tasks \citep{sun-jiang-2019-contextual,xian-et-al-2021-wecnlp}. One of its major applications is in optical character recognition (OCR). Recently, the technique has also found utility in image caption editing, where computer vision and natural language processing intersect \citep{wang-et-al-2022-eccv,yuan-et-al-2023-acm}.
Despite its significance, in written text processing, the task of text denoising has often been overlooked due to the perception that the provided datasets are already well-prepared. Prompting has proven successful in the text transfer task, even with small language models \citep{suzgun-etal-2022-prompt}. Therefore, we approach the processing of written text as a form of text transfer, employing prompting. 

In the text denoising experiment, our objective is to \textit{clean} the dataset used for automated essay scoring (AES) in preparation for the linear regression task. AES involves the use of an automated system to assign numerical grades or scores to essays written in a first language (L1) within an educational setting.
It is crucial to note that English AES systems rely on the ASAP dataset, initially provided in the Kaggle competition of 2012.\footnote{\url{https://www.kaggle.com/c/asap-aes}} This dataset has become widely adopted in various AES systems \citep{alikaniotis-alikaniotis-rei:2016:ACL,cummins-zhang-briscoe:2016:ACL,dong-zhang-yang:2017:CONLL,ridley-EtAl:2021:AAAI,chen-li-2023-pmaes,do-etal-2023-prompt}. 
The original ASAP dataset employed a named entity recognition approach using Stanford CoreNLP \citep{manning-EtAl:2014:P14-5} and a range of pattern matching rules to eliminate personally identifying information from the essays. Consequently, entities within the text are identified and replaced with strings initiated with the `@' symbol, such as \texttt{@PERSON1}.
Furthermore, more than 5\% of sentences exhibit encoding issues where UTF-8 symbols are not correctly displayed, in addition to the presence of non-word entities.

We classify sentences containing encoding issues and non-word entities as noise, and subsequently, undergo a denoising process to address them. Our hypothesis posits that enhancing text quality through denoising will yield improved linear regression results in AES. 
Throughout this paper, we use the term \textbf{prompt} in two distinct contexts:
(1) In the context of essay writing, a \textit{prompt} refers to a specific question, topic, or statement serving as the starting point for the essay.
(2) In the context of generative pre-trained transformers, a \textit{prompt} denotes the input or stimulus provided to the model to generate a response.

\section{Text denoising} \label{text-denoising}

In the context of the ASAP dataset, Figure~\ref{denoising-example} illustrates an instance of text denoising. For this process, we employ two prompts with \texttt{gpt-3.5-turbo-instruct}: one to address encoding errors and another to replace non-word entities with arbitrary entity names sequentially.\footnote{The prompting process costs \$6.87 for the entire 12,978 ASAP data entries, where we select sentences containing only such encoding errors and non-word entities.}
It is noteworthy that \texttt{gpt-3.5-turbo-instruct} has a tendency to correct grammatical errors in the original text during sentence generation. To restore the original words, we utilize the \texttt{.m2} annotation generated by ERRANT \citep{bryant-felice-briscoe:2017:ACL}. This annotation delineates the modifications made in the original text, allowing us to retain only the symbols with corrected encoding and the replaced non-word entities.

\begin{figure}[!ht]
\centering
\resizebox{.9\textwidth}{!}{
\footnotesize{
\begin{tabularx}{\textwidth}{|r X|}\hline 
\textsc{Original text}: & 
\texttt{people get @CAPS2 addicted that they don<U+0092>t exercize and become obeast, ...}\\
\textsc{Cleaned text}: & 
\texttt{people get too addicted that they don't exercize and become obeast, ...}\\\hline 
\end{tabularx}
}
}
\caption{Example of text denoising: the Unicode symbol \texttt{U+0092} is replaced with \texttt{'}, and \texttt{@CAPS2} is substituted with an arbitrary word}
\label{denoising-example}
\end{figure}

\section{Experiments and Results}
After text denoising, we utilize \texttt{roberta-base} \citep{liu-et-al-2019-roberta} for the linear regression task to derive the overall score. The results of the linear regression by quadratic weighted kappa (QWK), along with perplexity values, are presented in Table~\ref{results} for both the original and cleaned texts. 
It's important to note that all linear regression evaluations are conducted on the original text, even when the model is trained using the cleaned text, ensuring a fair comparison. We employ prompt-based 8-fold cross-validation, considering ASAP's eight different prompt essay sets, and the results represent the average \textsc{overall} scores.

\begin{table}[!ht]
\centering
{
\begin{tabular}{r | cc  c }\hline
    & {Sentence} ppl & {Token} ppl &  {Regression}\\ \hline 
\textsc{Original text}& 532.571& 65.535 & 0.6047 \\ 
\textsc{Cleaned text} & 520.023& 51.097 & 0.6143 \\  \hline
    \end{tabular}
}
\caption{Experiment results, including perplexity and regression using QWK}
\label{results}
\end{table}

\section{Discussion and Conclusion}
The ASAP dataset is often regarded as a given, with prior research primarily concentrating on constructing a regression system.
However, making even subtle modifications through denoising to the dataset can yield improved results. 
In the previous automatic writing evaluation system for L2 writing, \citet{lim-song-park-2023} observed that attention did not focus on spelling errors. Rather, a pre-trained large language model can be associated with more proper words instead of spelling errors to predict results. 
Through the cleaning process of the training dataset, we confirm their finding that the refined training dataset exhibits enhanced learning capabilities, leading to more accurate result prediction. 

A closer examination of the prompt-by-prompt QWK results reveals insightful nuances. 
In most cases, the cleaned text surpasses the original text, underscoring the efficacy of the denoising process. 
While the effectiveness of text denoising is evident, the nuanced variations across prompts necessitate careful consideration in experimental design.
As we explore these avenues, we anticipate that refining the dataset and incorporating additional features will contribute to the ongoing improvement of automated essay scoring systems.
The limitation of our study still lies in the choice of employing a simple transformer for the regression task instead of building our own neural regression system. As a result, despite our inability to achieve state-of-the-art results for the ASAP dataset, as demonstrated by previous neural AES systems incorporating numerous linguistic features \citep{ridley-EtAl:2021:AAAI,chen-li-2023-pmaes,do-etal-2023-prompt}, our approach yields relatively good results with minimal effort.


\subsubsection*{URM Statement}
The authors acknowledge that both authors of this work meet the URM criteria of ICLR 2024 Tiny Papers Track.

\bibliography{references}
\bibliographystyle{iclr2023_conference_tinypaper}

\appendix
\section{Appendix}

\subsection{Restoring the original text from \texttt{gpt}-corrected text}

We use the following two prompts with \texttt{gpt-3.5-turbo-instruct} for text denoising:

{\scriptsize
\begin{enumerate}\setlength\itemsep{0em}
    \item \texttt{Copy the sentence by replacing utf8 encoding error characters into the correct ascii symbols}
    \item \texttt{Copy the sentence by replacing @words into the real words}
\end{enumerate}
}

After obtaining the text generated by \texttt{gpt}, the \texttt{errant\_parallel} tool produces the \texttt{.m2} file. The annotations in the \texttt{.m2} file reveal the results of the prompting process, which involves fixing encoding errors and replacing non-word entities.

{\scriptsize
\begin{verbatim}
S 682|1|2 I think this because mere cases of suicide are from online bullying, 
people get @CAPS2 addicted that they don<U+0092>t exercize and become obeast, 
and because it is bad for the environment.
A 0 1|||U:OTHER||||||REQUIRED|||-NONE-|||0
A 19 20|||R:VERB|||don't|||REQUIRED|||-NONE-|||0        (***)
A 20 21|||R:SPELL|||exercise|||REQUIRED|||-NONE-|||0
A 23 24|||R:OTHER|||obese,|||REQUIRED|||-NONE-|||0    
\end{verbatim}
\begin{verbatim}
S 682|1|2 I think this because mere cases of suicide are from online bullying, 
people get @CAPS2 addicted that they don't exercize and become obeast, and 
because it is bad for the environment.
A 0 1|||U:OTHER||||||REQUIRED|||-NONE-|||0
A 5 6|||R:ADJ|||many|||REQUIRED|||-NONE-|||0
A 10 11|||R:OTHER|||caused by|||REQUIRED|||-NONE-|||0
A 13 14|||R:NOUN|||individuals|||REQUIRED|||-NONE-|||0  
A 15 16|||R:OTHER|||too|||REQUIRED|||-NONE-|||0         (***)
A 20 21|||R:SPELL|||exercise|||REQUIRED|||-NONE-|||0
A 23 24|||R:OTHER|||obese,|||REQUIRED|||-NONE-|||0
A 28 29|||R:ADJ|||harmful|||REQUIRED|||-NONE-|||0
A 29 30|||R:PREP|||to|||REQUIRED|||-NONE-|||0
\end{verbatim}
} where \texttt{S 682|1|2} represents \texttt{source}, the ASAP id, the prompt number, and the sentence id, respectively. 
We retain only the encoding errors and replace non-word entities in \texttt{A}s (the annotations corrected by \texttt{gpt}) marked with \texttt{(***)}. 
ERRANT enables us to regenerate the corrected sentence using the corresponding annotation information: \textit{people get too addicted that they don't exercize and become obeast,} ... as in the original text, instead of \textit{people get too addicted that they don't exercise and become obese,} ... as \texttt{gpt} generated.  

\subsection{Detailed experiment results}
Detailed prompt-by-prompt results are presented in Table~\ref{results-details}, 
including the outcomes of encoding fixed without replacing non-word entities (\textsc{Encoding fixed}), the evaluation conducted on the corresponding dataset rather than using the original text (\textsc{Cleaned'}), 
and the evaluation conducted on the original text to ensure a fair comparison (\textsc{Cleaned}) as in Table~\ref{results}.
During training, we utilize the default values of the \texttt{Trainer} class,\footnote{\url{https://huggingface.co/docs/transformers/main_classes/trainer}} employing the RoBERTa base model.
We normalize the score to the range of 0 and 1, and  we multiply the results by 100 to calculate QWK using the standard evaluation script provided by \texttt{ets.org}.

\begin{table}[!ht]
    \centering
\resizebox{\textwidth}{!}{    
{\footnotesize
\begin{tabular}{c  | cccc cccc | c}\hline 
&  Prompt 1 & Prompt 2 & Prompt 3& Prompt 4& Prompt 5& Prompt 6& Prompt 7& Prompt 8 & \textsc{Average}\\ \hline
\textsc{Original} & 
0.6187 & 
0.6308 & 
0.6962 & 
0.6626 & 
0.7158          & 
0.5924          & 
0.4622          & 
0.4586          & 
0.6047 \\

\textsc{Encoding fixed} & 
0.6517 & 
0.5855 & 
0.6872 & 
0.6290 & 
0.7485 & 
0.6174 & 
0.4692 & 
0.4943 & 
0.6103 \\

\textsc{Cleaned'} & 
0.7193 & 
0.5414 & 
0.6906 & 
0.6454 & 
0.7374 & 
0.6187 & 
0.4441 & 
0.4442 & 
0.6052 \\ 

\textsc{Cleaned} & 
0.7344 & 
0.5485 & 
0.6978 & 
0.6540 & 
0.7416 & 
0.6156 & 
0.4637 & 
0.4589 & 
0.6143 \\
\hline 
    \end{tabular}
}
}
\caption{Prompt-by-prompt QWK results: \textit{e.g.} \texttt{prompt 1} represents that prompt 1 is used as a test set, and prompts 2-7 as a training set, and prompt 8 as a dev set.}
\label{results-details}
\end{table}

We present a comprehensive set of results utilizing various metrics in Table~\ref{results-metric-details}, including:
Quadratic Weighted Kappa (QWK), 
Kendall Rank Coefficient (KRC), 
Spearman Rank Correlation (SRC), 
Pearson Correlation Coefficient (PCC), 
Mean Squared Error (MSE), and 
Root-Mean-Square Deviation or Root-Mean-Square Error (RMSD). 
This diverse range of metrics provides a thorough evaluation of our results from different perspectives.

\begin{table}[!ht]
    \centering
{    
{\footnotesize
\begin{tabular}{c  | ccc ccc }\hline 
&  QWK & KRC & SRC & PCC & MSE & RMSD  \\ \hline
\textsc{Original} & 0.6047 & 0.5477 & 0.6867 & 0.7062 & 0.0295 & 0.1676 \\
\textsc{Encoding fixed} & 0.6103 & 0.5426 & 0.6810 & 0.6982 & 0.0288 & 0.1652 \\
\textsc{Cleaned all'} & 0.6052 & 0.5479 & 0.6869 & 0.7050 & 0.0296 & 0.1675 \\
\textsc{Cleaned all} & 0.6143 & 0.5432 & 0.6813 & 0.7016 & 0.0287 & 0.1645 \\
\hline 
    \end{tabular}
}
}
\caption{Detailed results using various metrics.}
\label{results-metric-details}
\end{table}

\end{document}